\documentclass[conference,9pt,letterpaper]{IEEEtran}
\IEEEoverridecommandlockouts
\usepackage{amsmath,amssymb}
\usepackage{graphicx}
\usepackage{booktabs,array}
\usepackage{etoolbox}
\usepackage{dblfloatfix}
\usepackage{cite}
\usepackage[hidelinks]{hyperref}
\graphicspath{{figures/}}
\newcommand{\method}{Ternix}
\hypersetup{
  pdftitle={Ternix: Learning and Deploying Ternary Language Models with Shared Scaling},
  pdfauthor={Guoxiang Xu, Guozhen Ji, Qi Sun, Cheng Zhuo},
  pdfsubject={Ternix preprint},
  pdfkeywords={Large language models, knowledge distillation, ternary language models, CPU inference, edge computing}
}

\patchcmd{\section}{0.7ex plus 1ex minus 0ex}{0.7ex}{}
  {\PackageError{Ternix}{Unexpected IEEEtran section spacing}{Check the supplied IEEEtran class.}}
\patchcmd{\subsection}{0.7ex plus .5ex minus 0ex}{0.7ex}{}
  {\PackageError{Ternix}{Unexpected IEEEtran subsection spacing}{Check the supplied IEEEtran class.}}
\makeatletter
\patchcmd{\@afterheading}{\clubpenalty \@M}{\clubpenalty \@clubpenalty}{}
  {\PackageError{Ternix}{Unexpected heading paragraph hook}{Check LaTeX heading definitions.}}
\makeatother

\title{Ternix: Learning and Deploying Ternary Language Models with Shared Scaling}
\author{%
  \IEEEauthorblockN{Guoxiang Xu\textsuperscript{1,2}, Guozhen Ji\textsuperscript{1,2}, Qi Sun\textsuperscript{1,2,*}, and Cheng Zhuo\textsuperscript{2,*}}
  \IEEEauthorblockA{\textsuperscript{1}ChipFlux \quad \textsuperscript{2}Zhejiang University}
  \thanks{*Corresponding authors: Qi Sun and Cheng Zhuo.}
  \thanks{This work has been submitted to the IEEE for possible publication.
  Copyright may be transferred without notice, after which this version may no longer be accessible.}
}
\begin{document}
\maketitle
\begin{abstract}
Demand for running large language models (LLMs) on edge devices has continued to grow recently. CPU-centric deployment extends on-device intelligence to hardware without dedicated GPUs, as CPUs are the most ubiquitous processors across these systems. We present Ternix-9B, which jointly learns ternary connections and shared group and source scales across all text parameters through online distillation from Qwen3.5-9B. To improve CPU inference efficiency, we combine compact weight encoding, table-lookup computation, and parallel scheduling. Evaluations across seven full benchmarks and nine x86 and ARM CPUs show that Ternix-9B achieves a 50.16\% mean score (76.9\% of the BF16 teacher) with a 2.53 GB CPU execution weight file, while decoding at 32.19 tokens/s on an Intel Core Ultra 5 laptop and 51.85 tokens/s on Apple M6.

\end{abstract}
\section{Introduction}
In recent years, improvements in CPU computing power have made it possible to deploy large language models (LLMs) on CPU platforms~\cite{shen2023cpu}. Meanwhile, advances in LLMs have driven growing demand for everyday applications such as intelligent assistants and writing assistance~\cite{mobilellm2024,apple2024}. Since these applications often involve personal information, enabling models to perform inference locally on edge devices without uploading data to the cloud is particularly important for data protection and personal privacy. Moreover, local inference does not depend on network connectivity and offers additional advantages such as low latency and reduced variability. CPUs are widely available in edge settings, including personal computers, mobile devices, and embedded systems. Deploying LLMs on CPUs and improving CPU inference efficiency are therefore important ways to enable on-device intelligence~\cite{tmac2025}.

Hardware--software co-design is essential for LLM inference on CPUs~\cite{chen2024agile,polyq2026}. In model design, the BitNet family~\cite{bitnet2023,bitnet1582024} uses binary or ternary linear-layer weights, while Bitnet.cpp~\cite{bitnetcpp2025} combines compact encoding with specialized CPU kernels to reduce model storage and inference costs. On the hardware side, T-SAR~\cite{tsar2025} improves execution efficiency through SIMD hardware modifications tailored to ternary computation. For deployment, T-MAC~\cite{tmac2025} accelerates low-bit matrix operations on existing CPUs through table lookup. Sandwich~\cite{sandwich2026} separately optimizes execution plans and core allocation for prefill and decoding, and switches between these plans at runtime to improve CPU inference efficiency.

\begin{figure*}[!t]
  \centering
  \includegraphics[width=0.90\textwidth]{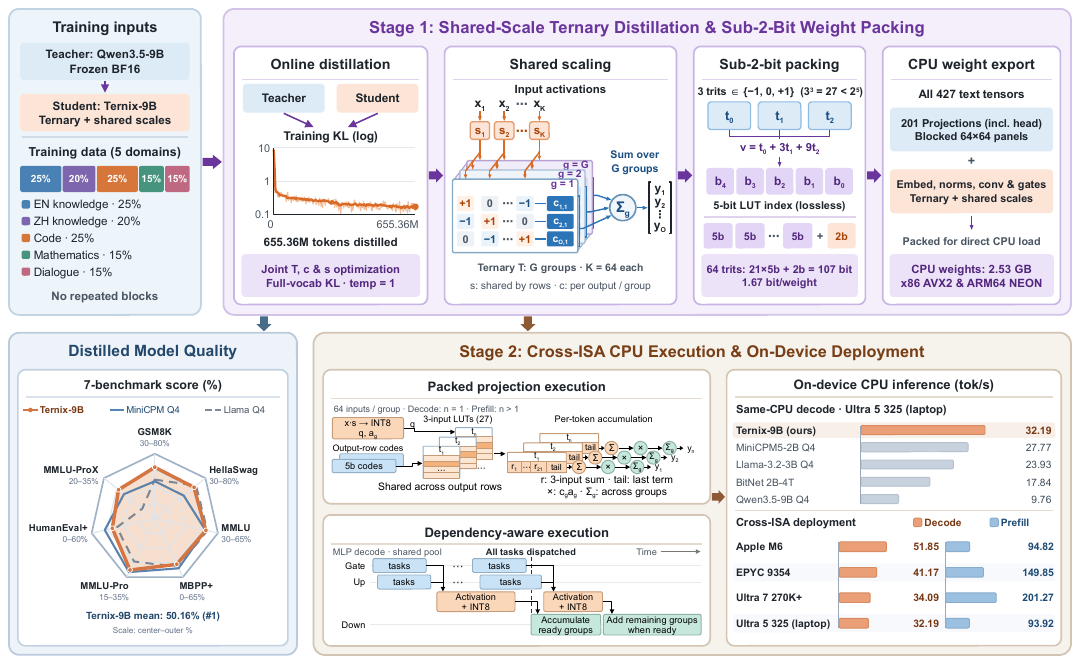}
  \caption{Overview of \method{}. Online distillation jointly learns ternary connections and shared scales; the plotted Ternix benchmark scores use GPU A8 evaluation. The learned representation is packed into cache panels for phase-specific lookup computation on existing x86 and ARM CPUs.}
  \label{fig:overview}
\end{figure*}

Although ternary representations substantially reduce model storage requirements, preserving model capabilities while achieving efficient CPU inference remains challenging. Specifically, directly restricting weights to three values can discard magnitude information and cause substantial capability degradation~\cite{catq2026}, while introducing additional scaling factors increases storage and computational overhead. Efficient CPU inference also requires compact, computation-friendly weight encoding~\cite{bitnetcpp2025}, data layouts and execution paths tailored to different instruction set architectures (ISAs)~\cite{tmac2025}, and careful scheduling of task dependencies and multithreaded pipelines. To address these challenges, we present \method{} (Fig.~\ref{fig:overview}). We first combine group scaling with source scaling shared across outputs to improve the expressiveness of ternary models with very few additional parameters. We then integrate this representation with compact encoding, direct table-lookup computation, and multithreaded pipelined execution to reduce storage costs and improve CPU inference efficiency. Experimental results demonstrate that \method{} is competitive in both model capability and on-device CPU inference speed.

Our main contributions are as follows:
\begin{enumerate}
  \item We propose a shared-scaling ternary model architecture covering all text parameters. We combine group scaling with source scaling shared across outputs to enhance the model's ability to represent weight magnitudes using few additional parameters, and jointly learn ternary connections and scaling parameters through online teacher distillation.
  \item We design a CPU inference method for x86 and ARM that combines compact ternary encoding, SIMD table-lookup computation, and multithreaded pipelined execution. We optimize data access and task scheduling separately for prefill and decoding to improve CPU inference efficiency.
  \item We train and deploy \method{}-9B, evaluate its model capabilities on seven full benchmarks, and conduct measurements on multiple x86 and ARM CPUs. Decoding throughput reaches 32.19 and 51.85 tokens/s on the Intel Core Ultra 5 325 (laptop) and Apple M6, respectively, demonstrating the feasibility and practical performance of on-device deployment.
\end{enumerate}

\section{Related Work}
In model training, knowledge distillation uses a teacher model to guide student learning and help preserve capabilities in compact models. TAID~\cite{taid2025} initially sets learning targets close to the student's current outputs and gradually shifts them toward the teacher's outputs, making learning easier for the student. BitDistill~\cite{bitdistill2025} converts existing floating-point models into ternary models through continual pre-training and distillation while seeking to preserve their original capabilities.

For CPU execution, Bitnet.cpp~\cite{bitnetcpp2025} accelerates ternary models through compact encoding and table-lookup computation, while llama.cpp~\cite{llamacpp} supports inference for various low-bit models. \method{} combines model training with CPU execution: it jointly learns ternary connections and shared scales, and optimizes weight encoding, table-lookup computation, and multithreaded pipelines to balance model capability, weight-file size, and inference efficiency.

\begin{figure}[t]
  \centering
  \includegraphics[width=0.75\columnwidth]{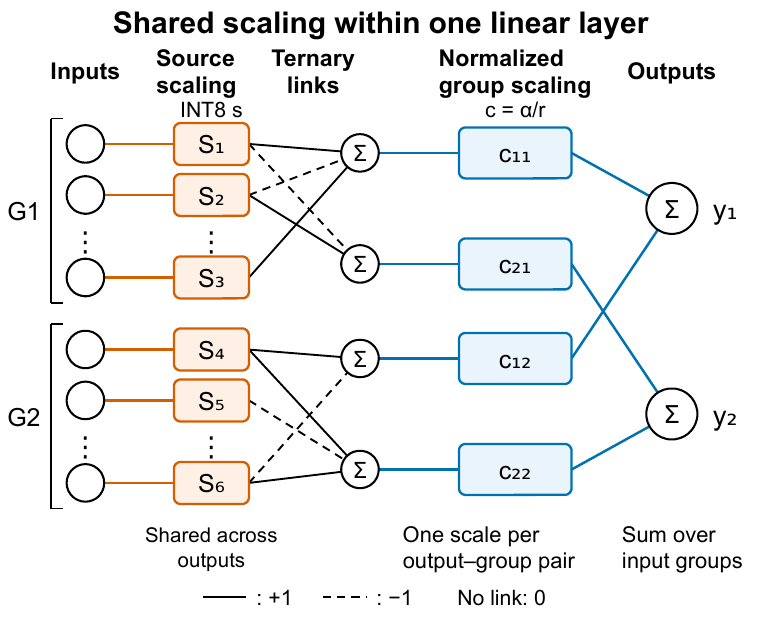}
  \caption{Shared scaling within a tensor. Source scales are reused across output rows; each group coefficient includes ternary RMS normalization. Solid, dashed, and absent links denote $+1$, $-1$, and $0$.}
  \label{fig:shared-scaling}
\end{figure}

\section{Method}
\label{sec:method}
\method{} couples a shared-scale model representation with CPU kernels that reuse the same scale structure. Fig.~\ref{fig:overview} summarizes the path from joint learning to packed execution.

\subsection{Shared-Scaling Model and Training}
\label{sec:model}

\subsubsection{Model structure}
\method{}-9B retains the text architecture and tokenizer of Qwen3.5-9B: 32 blocks comprising 24 Gated DeltaNet blocks and eight full-attention blocks, with hidden width 4096 and MLP width 12,288. All 427 learned text tensors, including embeddings, the output head, normalization, convolutions, and gates, use ternary connections and shared scales. To handle different tensor shapes uniformly, we reshape each tensor into a matrix, treating vectors as a single row, and divide each row into groups of 64 connections along the input dimension. Rows shorter than 64 connections form a single group.

\subsubsection{Shared scaling}
We jointly learn group scaling and source scaling: group scaling controls the overall magnitude of each weight group, while source scaling adjusts the relative magnitudes of input positions within a group and is shared across all output rows.

Consider a weight matrix with $O$ output rows, $G$ input groups per row, and $K$ connections per group. Indices $o$ and $g$ identify an output row and an input group, while $k$ and $j$ index positions within a group. Let $T_{ogk}\in\{-1,0,1\}$ denote a ternary connection, and let $\alpha_{og}$ and $s_{gk}$ denote the group magnitude and source scale, respectively. Their values after encoding and decoding are denoted by $\widetilde\alpha_{og}$ and $\widetilde s_{gk}$.

We normalize the ternary connections by their root mean square (RMS) to control group magnitude separately from the fraction of nonzero connections. Let $r_{og}$ denote the RMS with a lower bound, $c_{og}$ the normalized group coefficient, and $\widehat W_{ogk}$ the represented weight:
\begin{equation}
 \begin{aligned}
 r_{og}&=\max\left(\sqrt{\frac{1}{K}\sum_{j=1}^{K}T_{ogj}^{2}},\;0.2\right),\\
 c_{og}&=\frac{\widetilde\alpha_{og}}{r_{og}},\\
 \widehat W_{ogk}&=c_{og}T_{ogk}\widetilde s_{gk}.
 \end{aligned}
  \label{eq:shared-weight}
\end{equation}
As illustrated in Fig.~\ref{fig:shared-scaling}, inputs are first multiplied by source scales, summed through ternary connections, and then multiplied by group coefficients. A matrix uses $O\times G$ group magnitudes and $G\times K$ source scales, with source scales shared across all output rows.

To keep group magnitudes positive, we introduce learnable parameters $a_{og}$ and set $\alpha_{og}=\exp(a_{og})$.

Source scales control relative magnitudes within a group, so we fix their RMS to one and let the group magnitude control the overall scale. Specifically, let $\xi_g\in\mathbb R^{K-1}$ be a learnable vector and $H\in\mathbb R^{K\times(K-1)}$ a fixed Helmert basis. Their product gives a zero-mean log-scale vector $h_g=H\xi_g$, whose $k$th entry is $h_{gk}$. We define
\begin{equation}
 s_{gk}=\frac{\exp(h_{gk})}
 {\sqrt{\frac{1}{K}\sum_{j=1}^{K}\exp(2h_{gj})}}.
 \label{eq:source-scale}
\end{equation}

Let $v$ denote any group magnitude or source scale, $q$ its INT8 code, and $b$ a shared INT16 exponent. Both scale families use the same encoding and decoding:
\begin{equation}
 \begin{aligned}
 q&=\operatorname{clip}_{[-128,127]}
 \left(\operatorname{round}(v2^{-b})-128\right),\\
 \widetilde v&=(q+128)2^b.
 \end{aligned}
 \label{eq:fixed-scales}
\end{equation}
Exponents are shared per output row for group magnitudes and per input group for source scales, initialized with 25\% magnitude headroom, and then held fixed. Training updates FP32 continuous parameters and uses decoded values $\widetilde v$ in forward computation from the first update.

\subsubsection{Joint online distillation}
We initialize the student from the official BF16 weights of Qwen3.5-9B and set source scales to one. Within each group, we divide the original weights by 1.4 times their RMS and clip the result to $[-1.5,1.5]$ to initialize the continuous connection parameters $z$. Connections take $+1$ when $z>0.5$, $-1$ when $z<-0.5$, and zero otherwise. Initial group magnitudes minimize the squared error between the ternary representation and the original weights.

During online distillation, the frozen teacher and the student read the same text and predict the next token. The teacher assigns a probability to each candidate token, and the student learns this distribution. We compute the loss only at positions whose next token is contained in the same text block.

Let $p_i^T$ and $p_i^S$ denote the teacher's and student's probability distributions at the $i$th prediction, and let $N$ be the number of predictions included in training. We measure their difference using the KL divergence $D_{\mathrm{KL}}$ and average it to obtain the training loss $\mathcal L$:
\begin{equation}
 \mathcal L=\frac{1}{N}\sum_{i=1}^{N}
 D_{\mathrm{KL}}\!\left(p_i^T\middle\|p_i^S\right).
 \label{eq:distillation}
\end{equation}
Training uses a straight-through estimator to approximate gradients through ternarization, while AdamW jointly updates the continuous connection parameters and both scale families. Gradient computation includes the effect of group normalization. After each update, ternary connections and INT8 scale codes are regenerated for the next forward pass. Training uses BF16 dense computation with FP32 optimizer states.

\subsection{CPU Deployment}
\label{sec:cpu}

\subsubsection{Compact weight storage and input preprocessing}
\begin{figure}[!t]
  \centering
  \includegraphics[width=0.98\columnwidth]{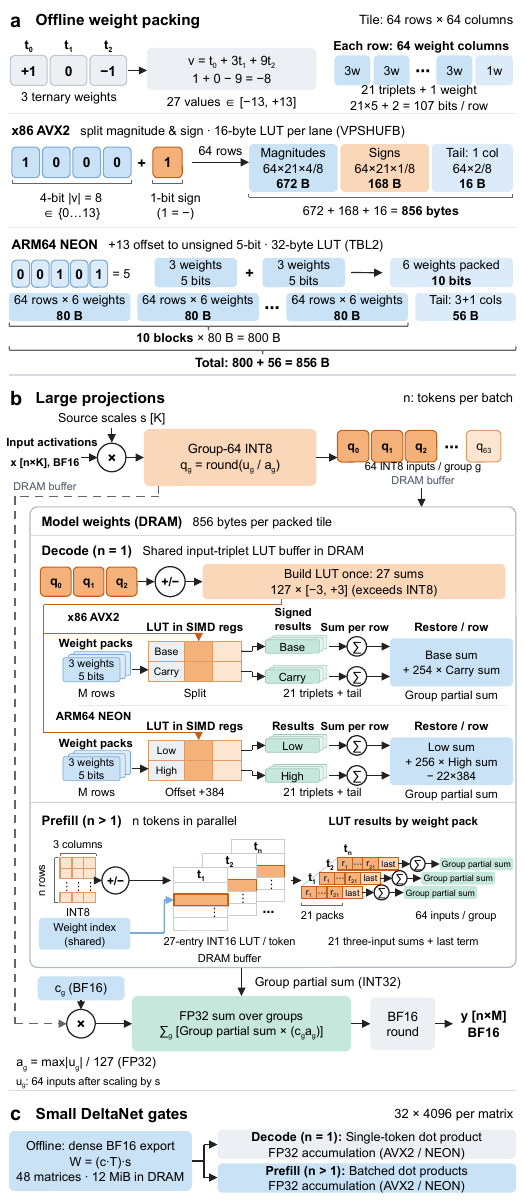}
  \caption{Packed projection execution on x86 and ARM. Input tables are built once per triplet and shared across output rows. Decode stacks denote output rows; each component accumulates 21 triplets and the tail per row. Prefill depth denotes tokens in both stacks; each result layer forms a 64-input group partial sum. Small gates use derived BF16 weights.}
  \label{fig:cpu-datapath}
\end{figure}

We compactly encode the weights of 201 large linear layers, including the language model output layer (LM head). Every three ternary connections use five bits, organized into $64\times64$ weight tiles for contiguous reads and SIMD table lookup. Including the remaining connection in each group, storage averages approximately 1.67 bits per connection. x86 and ARM use layouts adapted to AVX2 and NEON instructions, respectively (Fig.~\ref{fig:cpu-datapath}).

Inputs are first multiplied by source scales, then divided into groups of 64, scaled according to each group's maximum absolute value, and rounded to INT8. These prepared inputs are reused across all output rows of the same linear layer. We compute each group's integer partial sum, multiply it by the input scale and group coefficient, accumulate in group order using FP32, and convert the final result to BF16.

\subsubsection{Table-lookup computation for prefill and decoding}
For each triplet of INT8 inputs of a token, we precompute the 27 signed sums corresponding to ternary connection patterns and store them in a lookup table (LUT). All output rows share this table; each row uses its own weight codes to look up and accumulate results. This amortizes table construction across output rows and replaces per-connection multiplication with table lookup and addition.

Let $n$ denote the number of tokens processed together. Decoding ($n=1$) uses AVX2 or NEON byte table-lookup instructions. Results that exceed a single byte's range are stored in separate components and exactly restored after computation. In prefill ($n>1$), the same weight code selects a contiguous INT16 vector with one lookup result per token, and vector additions accumulate these results separately for each token.

\subsubsection{Dependency-aware parallel scheduling}
\begin{figure}[t]
  \centering
  \includegraphics[width=\columnwidth]{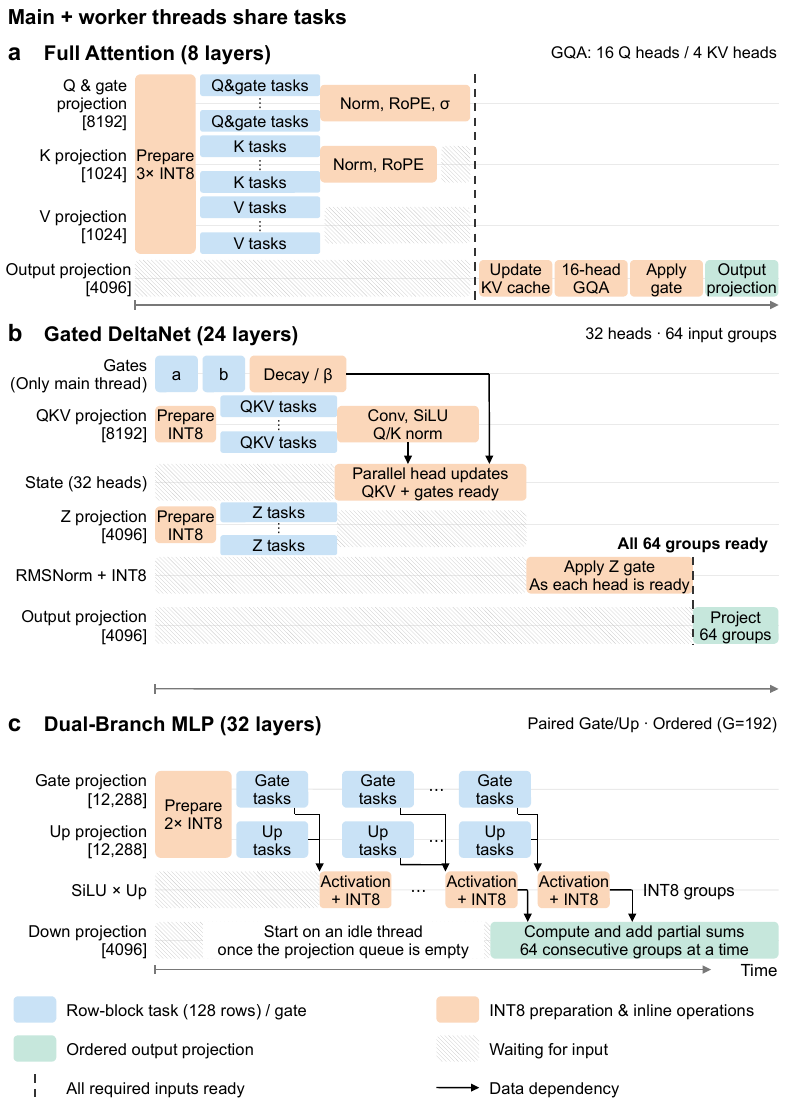}
  \caption{Dependency-aware execution within attention, DeltaNet, and MLP sub-layers. Logical task streams share a worker pool; downstream work starts when its required inputs are ready. Timelines are schematic, and layers and autoregressive steps remain sequential.}
  \label{fig:task-pipeline}
\end{figure}

A persistent worker pool partitions tasks by output rows and processes independent branches in parallel. Downstream operations start as soon as their required inputs are ready (Fig.~\ref{fig:task-pipeline}).

Full attention interleaves tasks for the combined query and gate projection and the key and value projections. Normalization and positional encoding run when the corresponding tasks finish; attention begins after the projection stage completes. In DeltaNet, the main thread computes the small gates while workers compute the query, key, value, and output-gating projections. Completed heads proceed through convolution, state updates, and gated normalization.

In the MLP, matching results from the gate projection (Gate) and up projection (Up) trigger SiLU activation of the Gate results and elementwise multiplication with the Up results. The resulting features feed the MLP output projection (Down), which maps intermediate features back to the model's hidden dimension.

Starting from the first input group, the MLP output projection (Down) accumulates ready groups in their original order while upstream tasks continue preparing the remaining inputs. Model layers and autoregressive steps remain sequential, while prefill and decoding use separate CPU core allocations.

Embeddings, normalization, convolution, and small gates use specialized kernels with coefficients derived from ternary connections and shared scales. Specifically, embeddings reconstruct only the requested token vectors, while normalization and convolution operate on reconstructed coefficients. To avoid table-lookup overhead for small matrices, we reconstruct the 48 small gate matrices as BF16 weights in advance for direct use during inference; these weights occupy 12\,MiB in total.

\section{Experiments and Results}
\label{sec:evaluation}
\subsection{Model Distillation and Multidimensional General Capability Benchmarks}
We use Qwen3.5-9B as the teacher to distill \method{}-9B on a mixture of English knowledge (25\%), Chinese knowledge (20\%), code (25\%), mathematics (15\%), and dialogue (15\%). Training runs for 40,000 steps with a batch size of eight and a sequence length of 2048. Peak learning rates for continuous connection parameters and scales are $3\times10^{-4}$ and $10^{-3}$, respectively, with 100-step warmup and cosine decay to 10\% of the peaks.

We evaluate the final model on seven benchmarks: GSM8K~\cite{gsm8k2021} for mathematical reasoning, HumanEval+ and MBPP+~\cite{evalplus2023} for code generation, HellaSwag~\cite{hellaswag2019} for commonsense reasoning, MMLU~\cite{mmlu2021} and MMLU-Pro~\cite{mmlupro2024} for general knowledge and reasoning, and MMLU-ProX (ZH)~\cite{mmluprox2025} for capabilities in Chinese. We compare with LFM3-1.7B~\cite{lfm3card2026}, MiniCPM5-2B~\cite{minicpm5card2026}, Llama-3.2-3B~\cite{llama32card2024}, and Qwen3.5-9B Q4\_K\_M~\cite{qwen35card2026}. The BF16 teacher serves as a reference.

GSM8K, MBPP+, and MMLU tasks use four, three, and five prompt examples, respectively; other tasks use none. Generation is greedy, capped at 256 tokens. Code tasks report pass@1, while other tasks report accuracy. MMLU uses answer-token scores, and HellaSwag uses byte-length-normalized likelihood scores. Large linear layers in both GPU and CPU implementations use source-scaled INT8 inputs in groups of 64 values. Weight-file sizes are in decimal GB; \method{}-9B uses its 2.53\,GB CPU execution file for both evaluation paths.

\begin{figure*}[!t]
  \centering
  \includegraphics[width=0.755\textwidth]{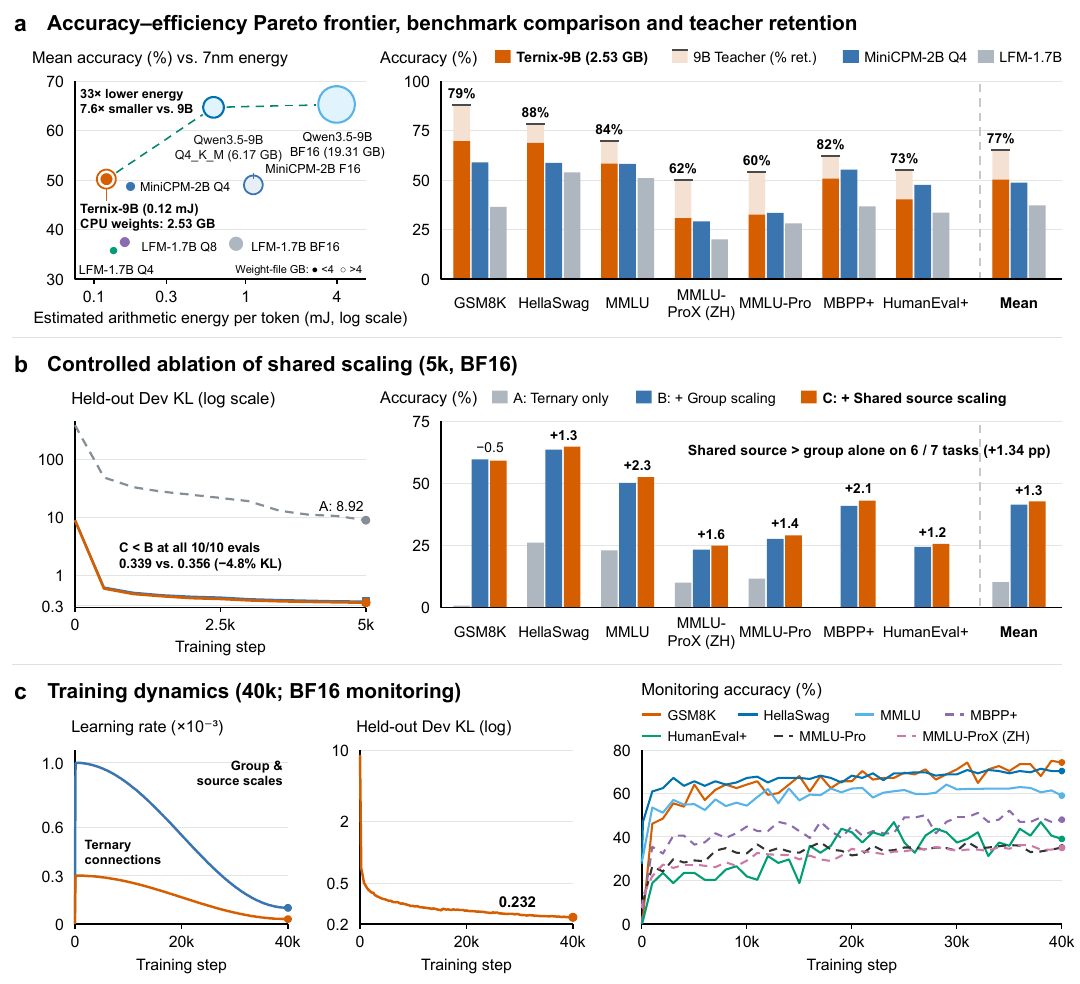}
  \caption{Model evaluation. (a) Full benchmarks, estimated projection energy, and weight-file size (bubble area); Ternix uses source-scaled INT8 inputs on GPU. (b) Shared-scaling ablation with reconstructed BF16 weights. (c) Training curves with BF16 subset monitoring.}
  \label{fig:learning-results}
\end{figure*}

\begin{table*}[!t]
\centering
\caption{Full benchmark results (\%). File denotes model weights. Best scores in the upper block are bold.}
\label{tab:main-benchmarks}
\footnotesize
\setlength{\abovecaptionskip}{3pt}
\setlength{\tabcolsep}{3pt}
\renewcommand{\arraystretch}{1.0}
\begin{tabular}{@{}llrrrrrrrrr@{}}
\toprule
Model & Precision & File (GB) & GSM8K & HumanEval+ & MBPP+ & MMLU-ProX (ZH) & HellaSwag & MMLU & MMLU-Pro & Mean \\
\midrule
LFM3-1.7B~\cite{lfm3card2026} & Q4\_K\_M & 1.10 & 32.45 & 32.93 & 34.13 & 19.85 & 52.78 & 50.43 & 27.47 & 35.72 \\
MiniCPM5-2B~\cite{minicpm5card2026} & Q4\_K\_M & 1.56 & 58.91 & 47.56 & 55.29 & 29.11 & 58.60 & 58.12 & 33.30 & 48.70 \\
LFM3-1.7B~\cite{lfm3card2026} & Q8\_0 & 1.85 & 37.91 & 34.15 & 37.30 & 19.85 & 53.96 & 50.82 & 28.05 & 37.43 \\
Llama-3.2-3B~\cite{llama32card2024} & Q4\_K\_M & 2.02 & 60.20 & 35.98 & 53.44 & 23.69 & \textbf{70.80} & 59.13 & 29.70 & 47.56 \\
\textbf{Ternix-9B (ours, GPU)} & Ternary & 2.53 & 69.67 & 40.24 & 50.79 & 30.71 & 68.76 & 58.36 & 32.59 & \textbf{50.16} \\
\textbf{Ternix-9B (ours, CPU)} & Ternary & 2.53 & \textbf{69.75} & 38.41 & 50.79 & \textbf{30.74} & 68.88 & 58.36 & 32.60 & 49.93 \\
LFM3-1.7B~\cite{lfm3card2026} & BF16 & 3.47 & 36.47 & 33.54 & 36.77 & 19.97 & 53.94 & 50.95 & 28.10 & 37.11 \\
MiniCPM5-2B~\cite{minicpm5card2026} & F16 & 5.04 & 53.60 & \textbf{51.22} & \textbf{56.08} & 29.89 & 59.24 & \textbf{59.14} & \textbf{33.81} & 49.00 \\
\midrule
Qwen3.5-9B~\cite{qwen35card2026} & Q4\_K\_M & 6.17 & 88.55 & 50.61 & 61.90 & 49.26 & 78.06 & 70.40 & 53.92 & 64.67 \\
Qwen3.5-9B~\cite{qwen35card2026} (teacher) & BF16 & 19.31 & 87.87 & 54.88 & 62.17 & 49.74 & 78.34 & 69.79 & 53.99 & 65.25 \\
\bottomrule
\end{tabular}
\end{table*}

To account for differences in execution order and floating-point rounding between CPU deployment and the GPU implementation, we separately test the CPU deployment computation path using native CPU evaluation and numerically validated GPU emulation. The CPU version largely preserves overall model capability.

Table~\ref{tab:main-benchmarks} summarizes the evaluation results. The Mean column is the equally weighted average of the seven task scores. Among the compared models under 4\,GB, \method{} has the highest mean score. Its GPU mean reaches 76.9\% of the teacher's.

Fig.~\ref{fig:learning-results}(a) compares model capability and theoretical energy consumption, estimated from projection operation counts and the unit energy costs of FP16/INT8 arithmetic~\cite{pokebnn2022}. Fig.~\ref{fig:learning-results}(c) shows the training curves.

\begin{table*}[!t]
\centering
\caption{Distillation comparison at 5,000 steps (\%, mean $\pm$ sample standard deviation over three seeds).}
\label{tab:distillation-baselines}
\footnotesize
\setlength{\abovecaptionskip}{3pt}
\setlength{\tabcolsep}{5pt}
\begin{tabular}{@{}lrrrrrrr@{}}
\toprule
Method & GSM8K & HumanEval+ & MBPP+ & MMLU-ProX (ZH) & HellaSwag & MMLU & MMLU-Pro \\
\midrule
TAID~\cite{taid2025} & $57.04\,\pm\,0.09$ & $25.61\,\pm\,2.11$ & $39.24\,\pm\,0.76$ & $23.64\,\pm\,0.45$ & $63.65\,\pm\,0.11$ & $50.21\,\pm\,0.87$ & $27.73\,\pm\,0.75$ \\
BitDistill~\cite{bitdistill2025} & $22.42\,\pm\,0.24$ & $15.24\,\pm\,1.83$ & $30.87\,\pm\,1.33$ & $11.64\,\pm\,1.29$ & $51.96\,\pm\,0.10$ & $30.16\,\pm\,1.95$ & $12.51\,\pm\,0.60$ \\
\textbf{Ternix-9B (ours)} & $\mathbf{61.16\,\pm\,1.69}$ & $\mathbf{25.81\,\pm\,0.35}$ & $\mathbf{43.74\,\pm\,0.66}$ & $\mathbf{24.68\,\pm\,0.20}$ & $\mathbf{64.76\,\pm\,0.18}$ & $\mathbf{53.28\,\pm\,0.98}$ & $\mathbf{29.12\,\pm\,0.08}$ \\
\bottomrule
\end{tabular}
\end{table*}

\subsection{Shared-Scaling Ablation}
We compare three variants: A uses only ternary connections, B adds group scaling, and C further adds source scaling shared across output rows. All three use the same data, teacher, and training settings for 5,000 steps and are evaluated using reconstructed BF16 weights. B and C start from identical reconstructed weights.

Fig.~\ref{fig:learning-results}(b) shows that B and C outperform A overall. Compared with B, C achieves lower held-out KL and higher scores on six of the seven benchmarks, indicating that shared source scaling further improves ternary model capability.

\subsection{Comparison with Distillation Baselines}
We compare \method{}, TAID~\cite{taid2025}, and BitDistill~\cite{bitdistill2025} on Qwen3.5-9B using identical data, 5,000 training steps, and three shared random seeds. Table~\ref{tab:distillation-baselines} reports means and sample standard deviations; \method{} achieves the highest mean score on all seven benchmarks.

\subsection{Comparison with CPU Inference Baselines}
We measure \method{}-9B inference throughput on nine x86 and ARM CPUs (Table~\ref{tab:cpu-platforms}). Prefill processes 1,024 input tokens and generates the first output token. Decode throughput is measured over the next 511 tokens.

\begin{table}[!t]
  \begingroup
    \centering
\caption{Final CPU deployment. D/P denotes decode/prefill threads; throughput is in token/s.}
\label{tab:cpu-platforms}
\footnotesize
\setlength{\tabcolsep}{2.25pt}
\begin{tabular*}{\columnwidth}{@{\extracolsep{\fill}}lrrr@{}}
\toprule
CPU & D/P & Decode $\uparrow$ & Prefill $\uparrow$ \\
\midrule
Apple M6 & 8/8 & 51.85 & 94.82 \\
EPYC 9354 (1 socket) & 16/16 & 41.17 & 149.85 \\
Core Ultra 9 285K & 16/24 & 38.55 & 238.66 \\
Apple M4 (laptop) & 6/6 & 35.91 & 58.19 \\
Ultra 7 270K Plus & 16/16 & 34.09 & 201.27 \\
Ultra 5 325 (laptop) & 6/8 & 32.19 & 93.92 \\
Xeon Gold 6348 (1 socket) & 16/16 & 31.16 & 137.86 \\
Core i9-14900HX (laptop) & 8/8 & 27.63 & 116.11 \\
Core i3-7100 & 2/2 & 6.84 & 25.29 \\
\bottomrule
\end{tabular*}

  \endgroup
  \par\smallskip
  \begingroup
    \centering
\caption{CPU runtimes on Ultra 5 325 (laptop). File and peak RAM in GB. Throughput (tok/s) and net power (W) are decode/prefill pairs from four runs. llama.cpp models use Q4\_K\_M.}
\label{tab:cpu-public}
\small
\setlength{\tabcolsep}{2pt}
\renewcommand{\arraystretch}{1.12}
\begin{tabular*}{\columnwidth}{@{\extracolsep{\fill}}llrrrr@{}}
\toprule
Model & Runtime & File & RAM & Throughput & Net power \\
\midrule
Ternix-9B & Ternix.cpp (ours) & 2.53 & 2.85 & 32.19/93.92 & 31.76/31.04 \\
BitNet 2B-4T & Bitnet.cpp~\cite{bitnetcpp2025} & 1.19 & 1.42 & 17.84/82.72 & 28.45/32.20 \\
MiniCPM5-2B & llama.cpp~\cite{llamacpp} & 1.56 & 2.60 & 27.77/100.08 & 24.83/27.71 \\
Llama-3.2-3B & llama.cpp~\cite{llamacpp} & 2.02 & 4.43 & 23.93/82.53 & 25.64/29.30 \\
Qwen3.5-9B & llama.cpp~\cite{llamacpp} & 6.17 & 11.61 & 9.76/37.85 & 26.03/30.23 \\
\bottomrule
\end{tabular*}

  \endgroup
\end{table}

We compare \method{}.cpp with Bitnet.cpp~\cite{bitnetcpp2025} and llama.cpp~\cite{llamacpp} on Ultra~5~325 (laptop), using six decode and eight prefill threads. Table~\ref{tab:cpu-public} reports model file size, peak memory, throughput, and net power. Net power is the CPU package power measured via Windows EMI minus mean idle power.

\method{}.cpp ranks first in decode throughput and second in prefill throughput. The results show efficient CPU deployment across platforms and leading performance against established CPU runtimes.

\begin{figure}[!t]
  \centering
  \includegraphics[width=\columnwidth]{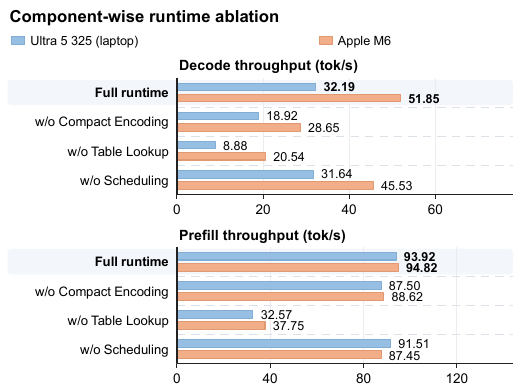}
  \caption{CPU runtime ablation. Full runtime values are from Table~\ref{tab:cpu-platforms}.}
  \label{fig:cpu-ablation}
\end{figure}

\subsection{CPU Deployment Ablation}
On Ultra~5~325 (laptop) and Apple M6, we remove one optimization at a time from the full runtime (Fig.~\ref{fig:cpu-ablation} and Table~\ref{tab:cpu-runtime-ablation}).

To ablate compact encoding, we replace each 5-bit triplet with three 2-bit weight codes. The runtime computes the corresponding lookup indices arithmetically while retaining the original weight tiling, lookup computation, and scheduling.

To ablate table lookup, we unpack the 5-bit weight triplets arithmetically and use SIMD integer multiply-accumulate in both prefill and decode. The weight format and scheduling are retained.

To ablate dependency-aware scheduling, we wait for all tasks in one stage to finish before starting the next. Tasks within each stage run in parallel with the same operators and thread configuration.

Removing table lookup causes the largest throughput drop in both phases on both CPUs. Compact encoding mainly improves decoding, and dependency-aware scheduling yields a larger gain on Apple M6.

\begin{table}[!t]
  \centering
  \caption{CPU runtime ablation (token/s). Each ablated variant is averaged over two runs.}
  \label{tab:cpu-runtime-ablation}
  \small
  \renewcommand{\arraystretch}{1.42}
  \setlength{\tabcolsep}{2.5pt}
  \begin{tabular*}{\columnwidth}{@{\extracolsep{\fill}}lrrrr@{}}
    \toprule
    & \multicolumn{2}{c}{Ultra 5 325 (laptop)} & \multicolumn{2}{c@{}}{Apple M6} \\
    \cmidrule(lr){2-3}\cmidrule(l){4-5}
    Runtime & Decode $\uparrow$ & Prefill $\uparrow$ & Decode $\uparrow$ & Prefill $\uparrow$ \\
    \midrule
    Full runtime & 32.19 & 93.92 & 51.85 & 94.82 \\
    w/o compact encoding & 18.92 & 87.50 & 28.65 & 88.62 \\
    w/o table lookup & 8.88 & 32.57 & 20.54 & 37.75 \\
    w/o scheduling & 31.64 & 91.51 & 45.53 & 87.45 \\
    \bottomrule
  \end{tabular*}
\end{table}

\section{Discussion}
\label{sec:discussion}
The current capabilities of \method{}-9B come from online teacher distillation, with room for further improvement. Future work could introduce reinforcement learning methods such as Group Relative Policy Optimization (GRPO)~\cite{deepseekmath2024}, using task feedback to further improve its general capabilities.

The small CPU weight file and on-device CPU deployment capability of \method{}-9B support local applications of specialized models. Through post-training for specific domains and tasks, \method{} could be adapted to chip design assistance~\cite{chipnemo2023} and on-device system agents~\cite{octopusv22024,apple2024}. Local inference avoids uploading raw data to the cloud, making it suitable for tasks involving sensitive information.

Current general-purpose processors offer limited direct support for ternary computation. Future work could develop dedicated chips for ternary connections and shared scaling, while exploiting the sparsity of zero connections to skip unnecessary computation and further improve inference speed and energy efficiency.

\section{Conclusion}
In summary, \method{} improves the expressiveness of ternary models through shared scaling and increases CPU inference efficiency by combining compact encoding, table-lookup computation, and parallel scheduling. Results from seven full benchmarks and deployment across multiple platforms show that \method{}-9B achieves competitive general capabilities with a small CPU weight file and runs efficiently on existing x86 and ARM CPUs, providing a viable solution for efficient LLM deployment on edge devices without GPUs or NPUs.

\clearpage
\IEEEtriggeratref{13}
\bibliographystyle{IEEEtran}
\bibliography{references}
\end{document}